%% file: 0_main.tex
\documentclass[letterpaper]{article} 
\usepackage{aaai25}  
\usepackage{times}  
\usepackage{helvet}  
\usepackage{courier}  
\usepackage[hyphens]{url}  
\usepackage{graphicx} 
\def\UrlFont{\rm}  
\usepackage{natbib}  
\usepackage{caption} 
\usepackage{algorithm}
\usepackage{algorithmic}

\usepackage{newfloat}
\usepackage{listings}
\DeclareCaptionStyle{ruled}{labelfont=normalfont,labelsep=colon,strut=off} 
\floatstyle{ruled}
\newfloat{listing}{tb}{lst}{}
\floatname{listing}{Listing}
\usepackage{pifont}
\usepackage{booktabs}
\usepackage{makecell}
\usepackage{color}
\usepackage{xcolor}
\usepackage{mathrsfs}
\usepackage{mathtools}
\usepackage[algo2e]{algorithm2e} 
\usepackage{paralist}
\usepackage{amssymb}
\usepackage{multirow}

\title{
\centering
GraphMoRE: Mitigating Topological Heterogeneity via \\
Mixture of Riemannian Experts}

\author{
    Zihao Guo\textsuperscript{\rm 1},
    Qingyun Sun\textsuperscript{\rm 1}\thanks{Corresponding author.},
    Haonan Yuan\textsuperscript{\rm 1},
    Xingcheng Fu\textsuperscript{\rm 2},
    Min Zhou\textsuperscript{\rm 3},
    Yisen Gao\textsuperscript{\rm 4},
    Jianxin Li\textsuperscript{\rm 1}
}

\affiliations{
    \textsuperscript{\rm 1}School of Computer Science and Engineering, Beihang University, China\\
    \textsuperscript{\rm 2}Key Lab of Education Blockchain and Intelligent Technology, Ministry of Education, Guangxi Normal University, China\\
    \textsuperscript{\rm 3}Huawei Technologies Co., Ltd, China\\
    \textsuperscript{\rm 4}Institute of Artificial Intelligence, Beihang University, China\\
    guozh@act.buaa.edu.cn, \{sunqy, yuanhn, gaoys, lijx\}@buaa.edu.cn, fuxc@gxnu.edu.cn, zhoumin27@huawei.com
}

\usepackage{bibentry}

\begin{document}

\maketitle

\newcommand{\MethodName}{GraphMoRE}
\begin{abstract}
Real-world graphs have inherently complex and diverse topological patterns, known as topological heterogeneity. Most existing works learn graph representation in a single constant curvature space that is insufficient to match the complex geometric shapes, resulting in low-quality embeddings with high distortion. This also constitutes a critical challenge for graph foundation models, which are expected to uniformly handle a wide variety of diverse graph data. Recent studies have indicated that product manifold gains the possibility to address topological heterogeneity. However, the product manifold is still homogeneous, which is inadequate and inflexible for representing the mixed heterogeneous topology. In this paper, we propose a novel \textbf{Graph} \textbf{M}ixture \textbf{o}f \textbf{R}iemannian \textbf{E}xperts (\textbf{\MethodName}) framework to effectively tackle topological heterogeneity by personalized fine-grained topology geometry pattern preservation. Specifically, to minimize the embedding distortion, we propose a topology-aware gating mechanism to select the optimal embedding space for each node. By fusing the outputs of diverse Riemannian experts with learned gating weights, we construct personalized mixed curvature spaces for nodes, effectively embedding the graph into a heterogeneous manifold with varying curvatures at different points. Furthermore, to fairly measure pairwise distances between different embedding spaces, we present a concise and effective alignment strategy. Extensive experiments on real-world and synthetic datasets demonstrate that our method achieves superior performance with lower distortion, highlighting its potential for modeling complex graphs with topological heterogeneity, and providing a novel architectural perspective for graph foundation models.

\end{abstract}

\input{1_introduction}
\input{2_related_work}
\input{3_preliminary}
\input{4_method}

\input{5_experiment}

\input{6_conclusion}

\newpage
\section*{Acknowledgements}
The corresponding author is Qingyun Sun. Authors of this paper are supported by the National Natural Science Foundation of China through grants No.62302023, No.62225202, and No.623B2010, the Beijing Natural Science Foundation QY24129, and the Fundamental Research Funds for the Central Universities. We owe sincere thanks to all authors for their valuable efforts and contributions.
\bibliography{aaai25}

\newpage
\clearpage
\appendix
\input{appendix/appendix.tex}

\end{document}

%% file: 1_introduction.tex
\section{Introduction}

Capturing the characteristics of structures with low distortion embeddings is critical to graph representation learning, while how to effectively improve the expressive capacity to adapt to topological heterogeneity remains an unresolved challenge. Particularly with the development of graph foundational models, there is an increasing necessity to uniformly process graph data from a wide variety of types and domains, imposing higher demands on adaptive feature extraction of complex structures.
Most existing graph representation learning methods~\cite{kipf2016semi,velivckovic2017graph,ying2021transformers,hou2024nc2d}, including existing works on graph foundation models\cite{liu2023graphprompt,sun2023all,zhao2024all,liu2023towards,maoposition}, embed graphs into Euclidean space, which are not conducive to capture the complex topological structures ~\cite{bronstein2017geometric,nickel2017poincare}.

In recent years, Riemannian representation learning has received wide attention due to the ability to naturally represent different topological structures~\cite{yang2024hypformer,fu2024hyperbolic}.
However, real-world graphs always exhibit topological heterogeneity, meaning they are hybrids of substructures with different topological characteristics, each suited to be represented in different curvature spaces. 
Most of previous works~\cite{liu2019hyperbolic,chami2019hyperbolic,bachmann2020constant,zhang2021lorentzian} exclusively consider a single constant curvature space, ignoring the problem of topological heterogeneity. 
Although some recent studies~\cite{wang2021mixed,sun2022self,cho2023curve} attempt to construct a mixed-curvature space (\textit{i.e.}, product manifold) that incorporates multiple Riemannian manifolds with different curvatures, they globally learn the representations of all nodes. 
In other words, the product manifold is homogeneous\cite{di2022heterogeneous} with the globally uniform curvature for each point in space, which is inadequate and inflexible for representing the mixed heterogeneous topology.

Considering the heterogeneity of topological patterns across different substructures of the graph, we aim to adaptively embed structures with different topological patterns in corresponding optimal embedding space, rather than applying the same curvature globally to the whole graph. 
To achieve automatic adaptation to heterogeneous topologies, there are two main issues need to be addressed. 
The first issue is about \textit{topology pattern identification}. The topological patterns of substructures in a graph are often not explicitly expressed and vary with resolution, making it difficult to directly divide the graph into subgraphs with different topological patterns and process them separately. 
The second issue is about \textit{optimal embedding space selection}. 
The appropriate type of Riemannian space and the optimal curvature vary from different topological properties. 
Moreover, the substructure of a graph is still complex and difficult to classify as a single type such as tree-like or cycle-like. 
Therefore, selecting the appropriate embedding space for different topological patterns also presents challenges.

\input{tables/intro_methods}


To address the first \textit{topology pattern identification} issue, we approach it from the perspective of local topology characterization, transforming the problem of finding substructures with different topological patterns into estimating the geometric properties around each node, thus providing a foundation for heterogeneity processing of the graph.
To address the second \textit{optimal embedding space selection} issue, we introduce Mixture-of-Experts (MoE) to model the complex geometric properties. 
In the MoE architecture, the experts can naturally correspond to different curvature spaces and construct personalized mixed curvature spaces for different nodes, while the gating module can adaptively route different nodes to the appropriate embedding space. 

To this end, we propose \MethodName, a Riemannian graph MoE to address the problem of topological heterogeneity. Specifically, we first propose a topology-aware gating mechanism, consisting of a multi-resolution local topology encoding module to estimate local geometric properties and a gating network which assigns Riemannian expert weights for nodes to route them to corresponding optimal embedding space. 
To achieve adaptive routing for nodes with different geometric properties, the gating network is guided by the goal of encouraging higher weights to be assigned to curvature spaces with lower distortion. 
Additionally, to adapt to complex geometric properties, we design diverse Riemannian experts considering the type of spaces and the degree of curving in space. 
The experts performs representation learning in differentiated curvature spaces to flexibly construct personalized mixed curvature spaces for nodes through expert weights.
Note that, our method is equivalent to the graph is embedded into a heterogeneous manifold with different curvatures at each point. 
Therefore, we provide an alignment strategy to measure the pairwise distance between nodes in different embedding spaces. 
We compare \MethodName~with relevant methods in Table \ref{intro_methods} and our method enjoys the advantage of learning fine-grained embeddings in heterogeneous manifold with diverse curvatures. 
Our contributions are as follows:
\begin{itemize}
    \item We propose analyzing topological heterogeneity from the perspective of local topology characterization, construct personalized embedding spaces for nodes, to minimize the distortion of heterogeneous topologies.
    \item To the best of our knowledge, we are the first to introduce the MoE into Riemannian representation learning, providing a new perspective for the design of graph foundation models based on Riemannian geometry.
    \item Extensive experiments demonstrate that our method outperforms advanced relevant methods, highlighting the potential of \MethodName~on topological heterogeneity.
\end{itemize}


%% file: tables/intro_methods.tex
\begin{table}[t]
\resizebox{\linewidth}{!}{
\begin{tabular}{lccc}
    \toprule
    \textbf{Method} & \makecell{\textbf{Curvature}\\\textbf{Diversity}} & \makecell{\textbf{Heterogeneous}\\\textbf{Manifold}} & \makecell{\textbf{Fine-grained}} \\
    \midrule
    HGCN  & \ding{55} & \ding{55} & \ding{55} \\
    $\kappa$-GCN & \ding{55} & \ding{55} & \ding{55} \\
    L-GCN & \ding{55} & \ding{55} & \ding{55} \\
    SELFMGNN & \ding{51} & \ding{55} & \ding{55} \\
    $\mathcal{Q}$-GCN & \ding{55} & \ding{51} & \ding{55} \\
    $\kappa$HGCN & \ding{55} & \ding{55} & \ding{51} \\
    MofitRGC & \ding{51} & \ding{51} & \ding{55} \\
    \midrule
    \textbf{\MethodName} & \ding{51} & \ding{51} & \ding{51} \\
    \bottomrule
\end{tabular}}
\caption{Comparison of \MethodName~and relevant methods.}
\label{intro_methods}
\end{table}

%% file: 2_related_work.tex
\section{Related Work}

\subsection{Riemannian Graph Learning}


Riemannian space has been introduced into graph representation learning and received wide attention~\cite{peng2021hyperbolic,yang2022hyperbolic}. 
To address the limitation of single constant curvature spaces in topological heterogeneity, $\mathcal{Q}$-GCN~\cite{xiong2022pseudo} introduces the pseudo-Riemannian manifold into GNNs. 
$\kappa$HGCN~\cite{yang2023kappahgcn} utilizes discrete curvature to improve message passing, alleviating inconsistency between local structure and global curvature. 
These works are orthogonal to ours.


In this paper, we mainly focus on the works related to mixed curvature space, which is mainly designed for non-uniform structured data~\cite{gu2018learning,skopek2019mixed}. 
DyERNIE~\cite{han2020dyernie}, M$^{2}$GNN~\cite{wang2021mixed} and AMCAD~\cite{xu2022amcad} respectively introduce product manifold into knowledge graph and retrieval system. SELFMGNN~\cite{sun2022self} proposes a mixed curvature graph contrastive learning, designs a hierarchical attention mechanism to fuse the representations of different curvature subspaces. FPS-T~\cite{cho2023curve} develops the graph transformer to mixed curvature space. 
MofitRGC~\cite{sun2024motif} introduces a diversified factor in the product manifold to provide flexibility. 
However, the product manifold proposed in most of these methods is still homogeneous. 
Although MofitRGC extends the original product manifold, its improvement is not intuitive and fundamental enough and is still insufficient to represent mixed heterogeneous topologies. 
The method proposed in this paper directly constructs personalized heterogeneous manifold for different geometric properties.

\subsection{Graph Mixture of Experts}

Mixture of Experts~\cite{jacobs1991adaptive,shazeer2017outrageously} aims at training multiple experts with different skills, widely used in fields such as large language models.
However, the exploration of MoE on graphs is still in the early stage. 
GraphDIVE~\cite{hu2022graphdive} and G-Fame++~\cite{liu2023fair} utilize MoE to learn diverse feature representations of graph, applying to imbalanced graph classification and fair graph representation learning. 
Considering the complex mixture of homophily and heterophily, GMoE~\cite{wang2024graph} and Node-MOE~\cite{han2024node} respectively propose using GNN with different hop numbers and different Laplacian filters as experts to enhance the ability to adapt to complex graph patterns. 
ToxExpert~\cite{kim2023learning} points out that a single GNN model cannot learn molecules with different structure patterns well and proposes to train multiple topology-specific expert models.
GraphMETRO~\cite{wu2024graphmetro} addresses out-of-distribution shifts by using a gating module to predict shift types and experts to generate shift-invariant representations. 
Note that, the diversity of experts and the design of the gating network are crucial. In this paper, we will mainly introduce how to design the components of MoE to mitigate the etopological heterogeneity.

%% file: 3_preliminary.tex
\section{Preliminary}
\label{sec:Preliminary}

\subsection{Riemannian Manifold}
\subsubsection{Manifold.}
A manifold is a generalization of high-dimensional surface, possessing the properties of Euclidean space locally. A Riemannian manifold is a smooth manifold coupled with a Riemannian metric $g$. In the tangent space $\mathcal{T}_{\mathbf{x}}\mathcal{M}^{d}$ of a point $\mathbf{x}$ on a d-dimensional Riemannian manifold $\mathcal{M}^{d}$, the Riemannian metric $g_{\mathbf{x}}$ defines a positive definite inner product $\mathcal{T}_{\mathbf{x}}\mathcal{M}^{d} \times \mathcal{T}_{\mathbf{x}}\mathcal{M}^{d} \rightarrow \mathbb{R}$, which is used to define the geometric properties and operations of the Riemannian manifold. The transformation of vector $\boldsymbol{v}$ between manifold $\mathcal{M}^{d}$ and tangent space $\mathcal{T}_{\mathbf{x}}\mathcal{M}^{d}$ is performed through logarithmic mapping $\log_{\mathbf{x}} (\boldsymbol{v}): \mathcal{M}^{d} \rightarrow \mathcal{T}_{\mathbf{x}}\mathcal{M}^{d}$ and exponential mapping $\exp_{\mathbf{x}} (\boldsymbol{v}): \mathcal{T}_{\mathbf{x}}\mathcal{M}^{d} \rightarrow \mathcal{M}^{d}$ at point $\mathbf{x}$. In Riemannian space, GNNs generally map nodes to the tangent plane through $\log_{\mathbf{0}} (\cdot)$ for message passing and aggregation, and then map nodes back to Riemannian space through $\exp_{\mathbf{0}} (\cdot)$, where $\mathbf{0}$ is the reference point.

\subsubsection{Constant Curvature Spaces.}
Curvature is a geometric quantity that describes the degree of curving in space. In the Riemannian manifold, the sectional curvature $\kappa(\mathbf{x})$ at each point $\mathbf{x}$ is defined. For a manifold with equal sectional curvature at all points, it is defined as a constant curvature space $\mathcal{M}^{d}_{\kappa}$. According to the sign of curvature $\kappa$, constant curvature spaces can be divided into three categories: spherical space ($\kappa > 0$), hyperbolic space ($\kappa < 0$), and Euclidean space ($\kappa = 0$), which are suitable for modeling data with different topologies respectively.

\subsection{Problem Formulation}
\subsubsection{Notations.}
A graph can be described as $\mathcal{G}=(\mathbf{A},\mathbf{X})$, where $\mathbf{A} \in \{0,1\}^{N \times N}$ is the adjacency matrix, $\mathbf{X} \in \mathbb{R}^{N \times d}$ is the node feature matrix, $N$ denotes the number of nodes and $d$ denotes the dimension of node feature. Let ${\mathcal{V}}$, ${\mathcal{E}}$ be the set of nodes and edges in the graph, respectively.

\subsubsection{Problem Definition.}
We primarily consider the problem of Riemannian representation learning on the graph with topological heterogeneity. Given a graph $\mathcal{G}=(\mathbf{A},\mathbf{X})$, in which various topological patterns exist. Our goal is to learn an encoding function $\psi: \mathcal{V} \rightarrow \mathcal{P}$, which captures the geometric properties around node $v$ and adaptively maps node $v$ to the personalized embedding space $\mathcal{P}$, where $\mathcal{P}$ is composed of diverse curvature spaces. Unlike existing work on mixed curvature, we embed different topological patterns into the corresponding optimal curvature spaces to maximize the preservation of the graph structure, rather than embedding all nodes into the same curvature space.

%% file: 4_method.tex
\section{\MethodName}

\begin{figure*}[t]
\centering
\includegraphics[width=\textwidth]{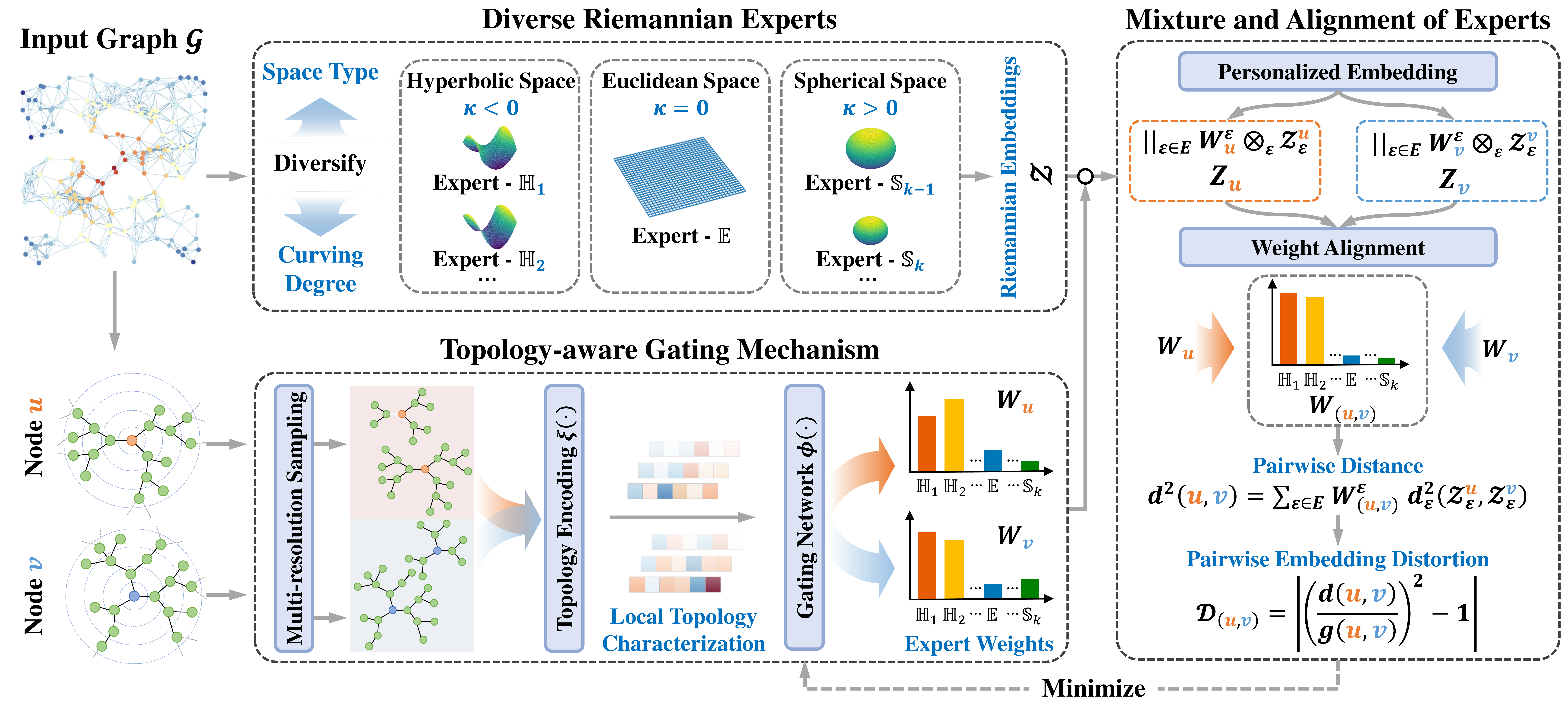} 
\caption{The overall framework of \textbf{\MethodName}. We first design diverse Rimannian experts in terms of both the type of curvature spaces and the degree of curving in space. Then we estimate the geometric properties around nodes from the perspective of local topology characterization, embed nodes into personalized mixed curvature spaces by fusing diverse Riemannian experts with the goal of minimizing embedding distortion. Finally, we measure pairwise distances through alignment strategy.}
\label{Framework}
\end{figure*}

In this section, we present \MethodName, a novel Graph Mixture of Riemannian Experts framework to address the problem of topological heterogeneity. The key insight is that we decompose topological heterogeneity from the perspective of local topology characterization. In brief, we first introduce diverse Riemannian experts~(Sec. \ref{sec:experts}) to provide a foundation for flexibly constructing diverse embedding spaces. Then, we design a topology-aware gating mechanism~(Sec. \ref{sec:gating}) to capture geometric properties around nodes and construct personalized mixed curvature spaces for each node~(Sec. \ref{sec:mixture}). Overall framework is shown in Figure \ref{Framework}.



\subsection{Diverse Riemannian Experts}
\label{sec:experts}
To perform heterogeneity processing on different topological patterns, we aim to provide personalized embedding spaces. However, the substructures of the graph are still complex, such that a single constant curvature space is not sufficient to match their geometric shapes well. Therefore, we consider multiple Riemannian spaces with different curvatures to better model local topology. In our method, we design each expert model as the Riemannian GNN in different curvature spaces. More importantly, we consider the diversity of Riemannian experts and provide personalized mixed curvature spaces through the gating network.


\subsubsection{Riemannian GNN.}
Before discussing further, we briefly introduce the operations of Riemannian expert with the backbone of the $\kappa$-stereographic model. Benefit from the $\kappa$-stereographic model provides a unified framework to describe manifolds with positive, negative, and zero curvature, our Riemannian experts learn embeddings in the unified space $\mathbb{U}_{\kappa}$. Specifically, $\kappa$-stereographic model is a smooth manifold $\mathcal{M}_{\kappa}^{d} = \{\mathbf{x} \in \mathbb{R}^d \mid-\kappa\|\mathbf{x}\|^2<1\}$. When $\kappa>0$, $\mathcal{M}_{\kappa}^{d}$ is the stereographic sphere model, while $\mathcal{M}_{\kappa}^{d}$ is the Poincaré ball model of radius $1/\sqrt{-\kappa}$ when $\kappa<0$.


The message passing and aggregation of Riemannian GNN in the $\kappa$-stereographic model can be formulated as:
\begin{align}
    &\mathbf{\hat{h}}_{u}^{(l+1)} = {\rm Agg} \left(\log_{\mathbf{0}}^{\kappa} \left(\mathbf{h}_{v}^{(l)}, v \in \mathcal{N}(u) \right) \right),\label{mpnn_agg}\\
    &\mathbf{h}_{u}^{(l+1)} = \exp_{\mathbf{0}}^{\kappa}  \left({\rm Comb} \left(\mathbf{h}_{u}^{(l)}, \mathbf{\hat{h}}_{u}^{(l+1)} \right) \right),\label{mpnn_comb}
\end{align}
where $\log_{\mathbf{0}}^{\kappa}$ and $\exp_{\mathbf{0}}^{\kappa}$ are exponential and logarithmic maps~\cite{bachmann2020constant}, $\mathcal{N}(\cdot)$ denotes the neighbor nodes, ${\rm Agg(\cdot)}$ and ${\rm Comb(\cdot,\cdot)}$ are neighbor aggregation and combination operators.

\subsubsection{Diversity of Experts.}
To enhance the ability to model complex topological patterns, we consider the diversity of Riemannian experts in terms of both the type of curvature spaces and the degree of curving in space. 
For the type of curvature spaces, we divide Riemannian experts into three categories, focusing on hyperbolic space $\mathbb{H}$, spherical space $\mathbb{S}$, and Euclidean space $\mathbb{E}$ respectively, where Euclidean space is a special case of the $\kappa$-stereographic model with zero curvature. 
Considering that Riemannian GNN only slightly adjusts the curvature within a relatively limited range~\cite{fu2021ace}, the initial value of curvature is crucial, which means that even if the type of curvature space has been considered, it is still necessary to consider the degree of curving in space. Therefore, we assign differentiated values to each type of Riemannian experts to initialize their curvature. The diverse Riemannian experts are denoted as $\boldsymbol{E}=\{\boldsymbol{E}_{\mathbb{H}_{1}},\boldsymbol{E}_{\mathbb{H}_{2}},...,\boldsymbol{E}_{\mathbb{E}},...,\boldsymbol{E}_{\mathbb{S}_{K}} \}$, each corresponding to a Riemannian GNN in different curvature spaces. 
Thus, we can flexibly construct personalized embedding spaces by fusing the representations of diverse Riemannian experts with the outputs of the gating network, providing the ability to model different topological patterns.


\subsection{Topology-aware Gating Mechanism}
\label{sec:gating}
As mentioned before, the topological patterns of substructures are often not explicitly expressed. We consider approaching it from the perspective of local topology characterization, transforming the problem into estimating the geometric properties around nodes. Specifically, we design a topology-aware gating mechanism, which includes multi-resolution local topology encoding and distortion guided gating network, to capture the geometric properties around nodes and route them to personalized embedding spaces.

\subsubsection{Multi-resolution Local Topology Encoding.}
To extract the characterization of the topology around nodes, we first consider local topology sampling centered on the nodes:
\begin{equation}
s_{v} = \operatorname{Sampler}(v,r),
\label{sample}
\end{equation}
where $s_{v}$ represents the subgraph sampled around the node $v$, $\operatorname{Sampler}(\cdot,\cdot)$ denotes the sampling strategy that can be flexibly chosen, $r$ denotes the scale of sampling. Considering that the local geometric properties exhibit some variation with changes in resolution, to obtain more abundant local topological characterizations, we further perform multi-resolution local topology sampling by Eq. \eqref{sample} as follows:
\begin{equation}
\mathcal{S}_{v} = \{\operatorname{Sampler}(v,r),r\in\mathcal{R} \},
\label{multi-sample}
\end{equation}
where $\mathcal{S}_{v}$ represents the sampled local topology subgraphs and $\mathcal{R}$ denotes the scale set of sampling.

Then, we introduce a GNN $\boldsymbol{\xi}(\cdot)$ designed for encoding the local topology, which embeds and pools the sampled local topology subgraphs $\mathcal{S}_{v}$, and concatenates them to obtain the local topology characterizations of node $v$:
\begin{equation}
\boldsymbol{{\mathcal{T}}}_{v} = \Vert ~{\rm Pooling}(\boldsymbol{\xi} (s),s \in \mathcal{S}_{v}),
\label{topolopy-embedding}
\end{equation}
where $\Vert$ denotes concatenation operation and ${\rm Pooling(\cdot)}$ denotes the pooling function.

\subsubsection{Distortion Guided Gating Network.}
After encoding the local topology of all nodes, we estimate the local geometric properties around the nodes and select the optimal embedding space for them. Specifically, the gating network $\boldsymbol{\phi}(\cdot)$ receives the local topology characterizations $\boldsymbol{{\mathcal{T}}}$ and outputs the weights of different Riemannian experts for each node:
\begin{equation}
\mathbf{W}_{v} = \operatorname{Softmax}(\boldsymbol{\phi}(\boldsymbol{{\mathcal{T}}}_{v})),
\label{expert-weight}
\end{equation}
where $\mathbf{W}_{v}=\{\mathbf{W}_{v}^{\boldsymbol{\varepsilon}},\boldsymbol{\varepsilon} \in \boldsymbol{E}\}$ is a weight vector, with each element representing the weight value of the corresponding Riemannian expert for node $v$. The topology characterizations of the sampled local topology subgraphs are similar for neighboring nodes, which ensures that the gating network assigns them nearly identical expert weights, enabling them to be embedded into similar embedding spaces.


For specific topological patterns, the gating network is expected to assign higher weights to the appropriate Riemannian experts. For the example in Figure \ref{Framework}, the topology around node $u$ is tree-like, so the output of the gating network is more biased towards hyperbolic Riemannian experts with negative curvatures, and experts with more matching curvatures are assigned relatively higher weights. 
To enable the gating network to have such ability, we consider embedding distortion as it can measure the inconsistency between the embedding distance of Riemannian experts and the graph distance. 
Specifically, for nodes in substructures with specific geometric properties, inappropriate expert weights cause these nodes to be embedded in mismatched curvature spaces, resulting in higher distortion. 
The optimization objective of minimizing the embedding distortion can promote the gating network to assign appropriate expert weights for nodes, which can be formalized as:
\begin{equation}
\mathcal{L}_{\mathcal{D}}=\frac{1}{\left|\mathcal{V}\right|^2} \sum_{i,j \in \mathcal{V}} \left|\left(\frac{d(i,j)}{g(i,j)}\right)^2-1\right|,
\label{}
\end{equation}
where $g(i,j)$ is the shortest path distance between node $i$ and $j$ on the graph, $d(i,j)$ is the distance between them in the Riemannian embedding space constructed in our method, which is related to the mixture and alignment of experts and will be described in detail in the next section.

\subsection{Mixture and Alignment of Experts}
\label{sec:mixture}
To construct a personalized embedding space for each node, we consider fusing the output of Riemannian experts with the expert weights which are assigned based on the local geometric properties, thus enabling heterogeneity processing of different nodes.
The output of Riemannian experts is denoted as $\boldsymbol{\mathcal{Z}}=\{\boldsymbol{\mathcal{Z}}_{\boldsymbol{\varepsilon}},\boldsymbol{\varepsilon} \in \boldsymbol{E} \}$, and the embedding of node $v$ can be formulated as:
\begin{equation}
\mathbf{Z}_{v} = \Vert_{\boldsymbol{\varepsilon} \in \boldsymbol{E}} \mathbf{W}_{v}^{\boldsymbol{\varepsilon} } \otimes_{\boldsymbol{\varepsilon} } \boldsymbol{\mathcal{Z}}_{\boldsymbol{\varepsilon} }^{v},
\label{}
\end{equation}
where $\boldsymbol{\mathcal{Z}}_{\boldsymbol{\varepsilon}}^{v}$ represents the embedding representation of node $v$ in the curvature space of expert $\boldsymbol{\varepsilon}$, $\otimes_{\boldsymbol{\varepsilon}}$ denotes the product operation defined on the manifold of expert $\boldsymbol{\varepsilon}$, that is, $w \otimes_{\boldsymbol{\varepsilon}} \mathbf{x}=\exp _{\mathbf{0}}^{\boldsymbol{\varepsilon}}\left(w \log_{\mathbf{0}}^{\boldsymbol{\varepsilon}}(\mathbf{x})\right)$. After the mixture of Riemannian experts, each node is embedded into a personalized mixed curvature space, which is equivalent to the graph is embedded into a heterogeneous manifold with varying curvatures at different points, thereby maximizing the preservation of mixed heterogeneous topological structures.

Moreover, due to the different spaces in which nodes are embedded, directly calculating the pairwise distance between nodes results in deviation. To achieve concise and effective embedding alignment, we consider integrating the expert weights of nodes $u$ and $v$ for each node pair $(u, v)$:
\begin{equation}
\mathbf{W}_{(u,v)} = \operatorname{Softmax}(\mathbf{W}_{u} \cdot \mathbf{W}_{v}),
\label{align}
\end{equation}
where $\mathbf{W}_{(u,v)}$ is the aligned expert weight. We remap node $u$ and $v$ through $\mathbf{W}_{(u,v)}$ to an aligned embedding space. The aligned embedding space is relatively advantageous for both when the original embedding spaces of two nodes are similar. For example, in Figure \ref{Framework}, the weights of nodes $u$ and $v$ are biased towards hyperbolic experts, and the aligned weight $\mathbf{W}_{(u,v)}$ can highlight experts which are important for both while reducing the weights of other experts. Otherwise, the aligned embedding space is neutral between the original embedding spaces of both. The embedding distance between nodes can be formulated as:
\begin{equation}
d^{2}(u,v) = \sum_{\boldsymbol{\varepsilon} \in \boldsymbol{E}} \mathbf{W}_{(u,v)}^{\boldsymbol{\varepsilon}} d^{2}_{\boldsymbol{\varepsilon}}(\boldsymbol{\mathcal{Z}}_{\boldsymbol{\varepsilon}}^{u},\boldsymbol{\mathcal{Z}}_{\boldsymbol{\varepsilon}}^{v}).
\label{embedding-distance}
\end{equation}

\subsection{Optimization Objective}
The overall optimization objective consists of the downstream task objectives and minimization of embedding distortion, which can be formulated as:
\begin{equation}
\mathcal{L} = \mathcal{L}_{task} + \lambda \mathcal{L}_{\mathcal{D}},
\label{loss}
\end{equation}
where $\lambda$ is a trade-off hyperparameter. The overall process of our method is summarized in Algorithm \ref{alg}. 

\subsection{Comlexity Analysis}
The overall time complexity is $O(\vert \mathcal{R} \vert \vert \mathcal{D} \vert + \vert \mathcal{D}_{s} \vert + K \vert \mathcal{D} \vert)$. Separately, the complexity of local topology sampling is $O(\vert \mathcal{R} \vert \vert \mathcal{D} \vert)$, and the complexity of the gating mechanism is $O(\vert \mathcal{D}_{s} \vert)$, where $\vert \mathcal{D} \vert$ is the size of the input graph and $\vert \mathcal{D}_{s} \vert$ is the size of the sampled subgraphs. The encoding process involves $K$ experts with the complexity of $O(K \vert \mathcal{D} \vert)$.

\begin{algorithm}[H] 
\caption{The overall training process of \MethodName}
\label{alg}
\DontPrintSemicolon
\KwIn{Graph $\mathcal{G}$; Number of experts $K$; Initial curvature set $C$; Number of training epochs $T$.} 
\KwOut{Predicted result of the downstream task.} 
Initialize Riemannian experts $\boldsymbol{E}$ with $C$; 

Sample local topology subgraphs $S$ for nodes by Eq. \eqref{multi-sample};

\For{$t = 1,2,...,T$}{

    \For{expert $\boldsymbol{\varepsilon}$ in $\boldsymbol{E}$}{
        Get the output $\boldsymbol{\mathcal{Z}}_{\boldsymbol{\varepsilon}}$ of expert $\boldsymbol{\varepsilon}$ by Eq. \eqref{mpnn_agg} and \eqref{mpnn_comb};
    }

    \tcp{Topology-aware Gating Mechanism}
    
    Calculate local topology characterizations $\boldsymbol{\mathcal{T}}$ by Eq. \eqref{topolopy-embedding};

    Get the expert weights $\mathbf{W}$ for each node by Eq. \eqref{expert-weight};

    \tcp{Mixture and Alignment of Experts}

    \For{each node pair $(u, v)$ in $\mathcal{G}$}{
        Get the aligned expert weights $\mathbf{W}_{(u,v)}$ by Eq. \eqref{align};
        
        Calculate embedding distance $d^{2}(u,v)$ by Eq. \eqref{embedding-distance};
    }

    \tcp{Update all parameters}
    Update model parameters by minimizing $\mathcal{L}$ by Eq. \eqref{loss};
  
}

\end{algorithm}

%% file: 5_experiment.tex
\section{Experiment}
To evaluate \MethodName\footnote{Code is available at https://github.com/RingBDStack/GraphMoRE.} proposed in this paper, we conduct comprehensive experiments and further analyze the effectiveness of \MethodName~on topological heterogeneity.


\input{tables/exp_lp}
\input{tables/exp_nc_F1}

\subsection{Experimental setup}
\subsubsection{Datasets.}
\label{sec:exp_Datasets}
We conduct experiments on a variety of real-world datasets, including citation networks~(Cora~\cite{sen2008collective}, Citeseer~\cite{kipf2016semi}, PubMed~\cite{namata2012query}), airline networks~(Airport~\cite{chami2019hyperbolic}) and co-purchase networks~(Photo~\cite{shchur2018pitfalls}). In addition, we generate graphs with topological heterogeneity for further comparison. The statistics and topological heterogeneity analysis are detailed in the Appendix~\ref{sec:datasets}.

\subsubsection{Baselines.}
We choose a variety of baselines including Euclidean methods and Riemannian methods. GCN~\cite{kipf2016semi}, GAT~\cite{velivckovic2017graph}, SAGE~\cite{hamilton2017inductive} are representatives of Euclidean GNNs. HNN~\cite{ganea2018hyperbolic}, HGCN~\cite{chami2019hyperbolic}, $\kappa$-GCN~\cite{bachmann2020constant}, LGCN~\cite{zhang2021lorentzian}, $\mathcal{Q}$-GCN~\cite{xiong2022pseudo}, MofitRGC~\cite{sun2024motif} are competitive Riemannian methods, where $\mathcal{Q}$-GCN and MofitRGC consider the topological heterogeneity.


\subsubsection{Settings.}
\label{sec:setting}
We set the number of layers to 2 for all methods. For other model settings, we adopt the default values in the corresponding papers. All reported numbers are averaged over ten independent runs for fair comparison. 

\subsection{Performance Evaluation}

\subsubsection{Performance on Real-world Graphs.}
We evaluate our method on link prediction and node classification tasks, and the results are summarized in Table \ref{exp_lp} and Table \ref{exp_nc_F1}. 
For the link prediction task, we use Fermi-Dirac decoder~\cite{nickel2017poincare} where the distance between two nodes in our method is calculated by Eq.~\eqref{align} and~\eqref{embedding-distance}. As observed, \MethodName~achieves the best results on all datasets among 9 baselines, and compared with the runner-up method, we achieve gains of 2.01\%, 1.94\%, and 1.92\% respectively in AUC score on Cora, Citeseer, and PubMed.
For the node classification task, we use GCN, GAT, and SAGE as backbone respectively. \MethodName~achieves the best results on the vast majority of datasets.


\subsubsection{Performance on Synthetic Graphs.}
\begin{figure}[t]
\centering
\includegraphics[width=0.48\textwidth]{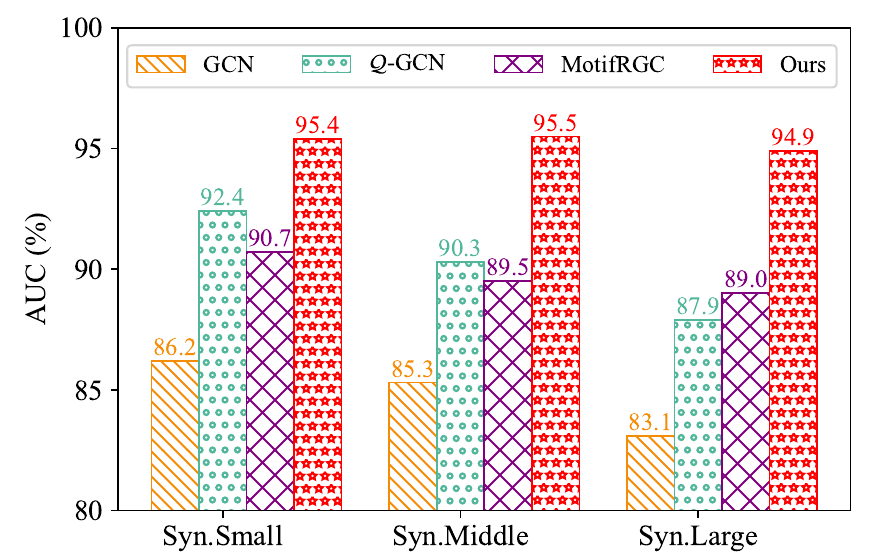} 
\caption{Performance comparison on synthetic graphs with topological heterogeneity evaluated by AUC (\%).}
\label{exp_synthetic}
\end{figure}

To further verify the ability to model mixed heterogeneous topologies, we evaluate our method on three different scale synthetic graphs with topological heterogeneity. We select the Riemannian methods $\mathcal{Q}$-GCN and MotifRGC which are designed for topological heterogeneity and the Euclidean method GCN, as baselines on the link prediction task. As shown in Figure \ref{exp_synthetic}, \MethodName~significantly outperforms all baselines. In addition, as the scale of datasets increases, the performance of all baselines declines to varying degrees, while \MethodName~consistently maintains excellent performance. This phenomenon is attributed to we analyze topological heterogeneity from the perspective of local topology characterization, embedding each node into corresponding personalized embedding space rather than modeling all nodes globally, which decouples \MethodName~from the scale of the dataset. 

\input{tables/exp_ablation}

\subsubsection{Ablation Study.}
We conduct ablation studies with four variants on two downstream tasks to verify the effectiveness of each component in \MethodName. Here, we use GCN as the classifier backbone for the node classification task. 
\begin{itemize}
    \item \textbf{\MethodName~(\textit{w/o distortion})}: we remove the embedding distortion $\mathcal{L}_{\mathcal{D}}$ from the optimization objective. 
    \item \textbf{\MethodName~(\textit{w/o gating})}: we replace the gating mechanism with an MLP. 
    \item \textbf{\MethodName~(\textit{w/o diverse})}: we initialize all Riemannian experts with the same curvature. 
    \item \textbf{\MethodName~(\textit{w/o align})}: we ignore Eq.~\eqref{align} and Eq.~\eqref{embedding-distance} and directly calculate the pairwise distance. 
\end{itemize}
The results are summarized in Table \ref{exp_ablation}, which shows that missing any component of \MethodName~leads to a degradation of the performance. We analyze the results and find that: 
\begin{inparaenum}[i)]
\item For \MethodName~(\textit{w/o distortion}) and \MethodName~(\textit{w/o gating}), there will be deviations in selecting the optimal embedding space for nodes due to the lack of embedding distortion guidance or local topology characterization.  
\item For \MethodName~(\textit{w/o diverse}), the performance degradation is most significant, which is because the lack of diversity among Riemannian experts limits the ability to flexibly construct personalized embedding spaces.  
\item For \MethodName~(\textit{w/o align}), directly calculating the pairwise distance between different embedding spaces leads to deviation.
\end{inparaenum}

\subsubsection{Comparison of Embedding Distortion.}
\begin{figure}[t]
\centering
\includegraphics[width=0.45\textwidth]{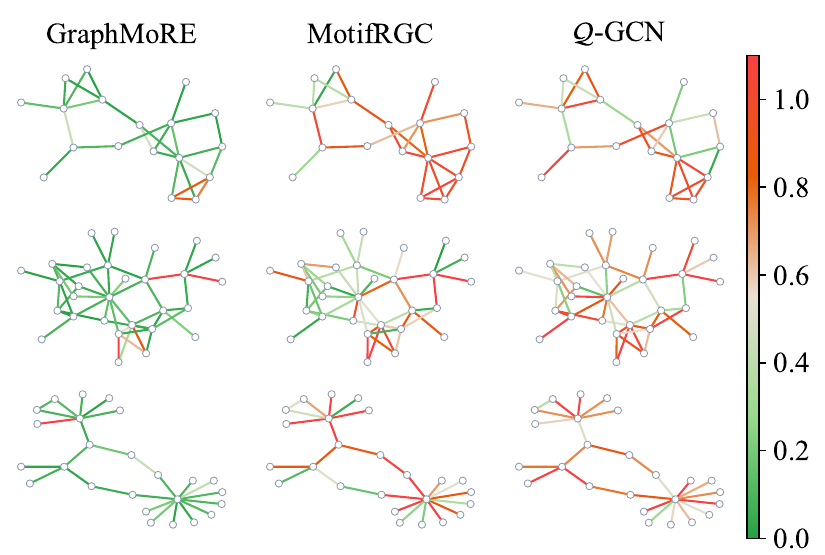} 
\caption{Visualization of distortion of different local topology structures in Cora. Closer to red, more distortion.}
\label{fig:local_distortion}
\end{figure}

Embedding distortion reflects the expressive ability for the graph structure, and lower embedding distortion indicates better preservation of the graph structure. 
We evaluate the average embedding distortion of our method and baselines designed for topological heterogeneity on the link prediction task and report the results in Table \ref{table:avg_distortion}. 
Obviously, \MethodName~has the lowest average embedding distortion on all datasets, which benefits from we consider local topologies with different geometric properties from the perspective of local topology characterization and construct personalized embedding spaces for them.
Additionly, we visualize some examples from Cora in Figure \ref{fig:local_distortion}, and it can be observed that \MethodName~has relatively low distortion for mixed heterogeneous topologies, indicating that \MethodName~has excellent expressive ability for topological heterogeneity. In contrast, methods that globally embed all nodes cannot effectively capture different geometric properties and process them differentially, resulting in higher embedding distortion.

\input{tables/avg_distortion}


%% file: tables/exp_lp.tex
\begin{table*}[htbp]
  \centering
    \resizebox{\linewidth}{!}{
    \begin{tabular}{lcccccccccc}
    \toprule
    \multirow{2}[2]{*}{\textbf{Method}}      & \multicolumn{2}{c}{\textbf{Cora}} & \multicolumn{2}{c}{\textbf{Citeseer}} & \multicolumn{2}{c}{\textbf{Airport}} & \multicolumn{2}{c}{\textbf{PubMed}} & \multicolumn{2}{c}{\textbf{Photo}}\\
    \cmidrule(r){2-3} \cmidrule(r){4-5} \cmidrule(r){6-7} \cmidrule(r){8-9} \cmidrule{10-11}
     & AUC   & AP    & AUC   & AP    & AUC   & AP    & AUC   & AP   & AUC   & AP\\
    \midrule
    GCN   & 91.88±0.33 & 92.44±0.36 & 90.10±0.65 & 90.66±0.57 & 91.95±0.91 & 91.35±0.69 & 93.79±2.91 & 93.49±2.73 & 92.17±2.57 & 90.42±2.71 \\
    GAT   & 90.63±0.30 & 91.13±0.31 & 88.83±0.64 & 88.81±0.18 & 89.55±1.22 & 89.25±1.34 & 88.87±1.57 & 87.96±1.30 & 94.06±0.40 & 93.15±0.50 \\
    SAGE  & 89.29±0.59 & 90.03±0.69 & 86.93±0.53 & 88.47±0.69 & 89.04±0.76 & 85.44±1.36 & 89.97±0.40 & 90.76±0.31 & 92.87±2.28 & 91.28±2.51 \\
    \midrule
    HNN   & 90.85±0.44 & 89.88±0.49 & 95.27±0.48 & 95.41±0.52 & 92.63±0.42 & 92.96±0.21 & 93.04±0.19 & 92.06±0.14 & 92.77±0.31 & 90.12±0.32 \\
    HGCN  & 93.62±0.25 & 93.73±0.27 & 94.87±0.41 & 95.28±0.37 & 93.50±0.36 & 94.13±0.28 & 95.20±0.15 & 95.16±0.16 & 96.96±0.93 & 96.12±0.94 \\
    $\kappa$-GCN & 93.81±1.95 & 94.97±1.55 & \underline{96.76±1.20} & \underline{97.39±0.88} & 93.87±0.34 & 93.18±0.23 & 97.17±0.12 & 97.05±0.13 & 97.29±0.09 & 96.72±0.12 \\
    LGCN  & 93.75±0.30 & 94.47±0.24 & 95.50±0.20 & 95.99±0.14 & 94.20±0.26 & 94.43±0.20 & 96.20±0.11 & 96.25±0.16 & \underline{97.50±1.10} & \underline{96.96±1.20} \\
    \midrule
    $\mathcal{Q}$-GCN & 93.66±0.17 & 92.96±0.12 & 94.41±0.25 & 93.93±0.15 & 94.99±0.33 & 94.49±0.21 & 95.21±0.05 & 95.14±0.11 & 97.33±0.04 & 96.53±0.06 \\
    MotifRGC & \underline{95.90±0.71} & \underline{96.51±0.54} & 96.04±0.39 & 96.33±0.43 & \underline{96.86±0.39} & \underline{96.65±0.28} & \underline{97.26±0.76} & \underline{97.20±0.78} & \multicolumn{2}{c}{OOM}\\
    \midrule
    \textbf{\MethodName}  & \textbf{97.91±0.10} & \textbf{98.42±0.09} & \textbf{98.70±0.20} & \textbf{98.84±0.14} & \textbf{97.53±0.28} & \textbf{96.71±0.36} & \textbf{99.18±0.05} & \textbf{99.22±0.05} &  \textbf{98.83±0.04} & \textbf{98.53±0.06}  \\
    \bottomrule
    \end{tabular}}
\caption{AUC and AP (\% ± standard deviation) of Link Prediction on real-world graph datasets. The best results are indicated in \textbf{bold} and the runner-ups are \underline{underlined}. OOM denotes Out-of-Memory.}  
  \label{exp_lp}%

\end{table*}%

%% file: tables/exp_nc_F1.tex
\begin{table*}[t]
  \centering
  \setlength{\tabcolsep}{5pt}
  \resizebox{\linewidth}{!}{
    \begin{tabular}{lcccccccccc}
    \toprule
     \multirow{2}[2]{*}{{\textbf{Method}}}     & \multicolumn{2}{c}{\textbf{Cora}} & \multicolumn{2}{c}{\textbf{Citeseer}} & \multicolumn{2}{c}{\textbf{Airport}} & \multicolumn{2}{c}{\textbf{PubMed}} & \multicolumn{2}{c}{\textbf{Photo}}\\
    \cmidrule(r){2-3} \cmidrule(r){4-5} \cmidrule(r){6-7} \cmidrule(r){8-9} \cmidrule{10-11}
     & W-F1  & M-F1  & W-F1  & M-F1  & W-F1  & M-F1  & W-F1  & M-F1  & W-F1  & M-F1 \\
    \midrule
    GCN   & 79.41±1.25 & 78.92±1.04 & 65.92±2.13 & 62.46±1.73 & 81.38±0.82 & 77.44±0.82 & 75.95±0.40 & 75.74±0.34  & 91.68±0.70 & 89.77±1.11 \\
    GAT   & 77.49±1.11 & 76.98±1.01 & 65.43±1.46 & 61.87±1.38 & 83.06±1.02 & 80.33±1.03 & 75.33±0.89 & 75.04±0.84  & 92.63±0.59 & 90.96±0.76 \\
    SAGE  & 77.47±0.71 & 76.61±0.87 & 64.03±1.11 & 60.24±1.56 & 84.07±2.19 & 81.27±2.17 & 73.97±0.90 & 73.90±0.87  & 88.09±1.56 & 85.25±1.90 \\
    \midrule
    HNN   & 58.98±0.52 & 57.15±0.70 & 59.52±0.51 & 57.38±0.68 & 70.71±1.47 & 54.49±1.81 & 69.56±0.71 & 68.85±0.49  & 90.90±0.42 & 89.34±0.53 \\
    HGCN  & 77.78±0.63 & 73.94±0.61 & 67.51±0.81 & 60.56±0.83 & 84.82±1.46 & 81.23±2.06 & 78.69±0.59 & \underline{77.78±0.50}  & 90.78±0.28 & 88.10±0.42 \\
    $\kappa$-GCN & 78.67±1.00 & 78.01±0.73 & 63.88±0.69 & 60.28±0.60 & 87.40±1.64 & 84.33±2.08 & 77.31±1.71 & 76.37±1.54  & 92.31±0.45 & 90.52±0.76 \\
    LGCN  & 80.19±0.98 & 79.05±0.81 & 67.94±1.78 & 61.41±4.25 & 89.03±1.24 & 84.50±2.00 & 77.19±0.41 & 76.92±0.44  & 92.97±0.32 & 90.53±1.23 \\
    \midrule
    $\mathcal{Q}$-GCN & 75.97±0.97 & 74.03±0.96 & 68.89±1.70 & 62.68±1.48 & 89.66±0.67 & 85.46±0.86 & \textbf{80.61±0.81} & \textbf{79.55±0.91}  & 91.85±0.47 & 90.30±0.63 \\
    MotifRGC & \underline{80.91±0.64} & \underline{80.19±0.63} & 68.31±1.41 & 64.63±1.39 & 84.53±2.32 & 83.96±2.38 & \underline{79.13±1.37} & 77.71±1.34  & \multicolumn{2}{c}{OOM}\\
    \midrule
    \textbf{\MethodName}$\rm _{GCN}$  & 80.51±0.91 & 79.44±0.77 & \textbf{69.73±0.70} & \textbf{65.98±0.67} & \underline{91.75±0.93} & \underline{91.49±0.98} & 77.16±0.61 & 76.50±0.76  & 93.18±0.31 & 91.86±0.34 \\
    \textbf{\MethodName}$\rm _{GAT}$ & 79.81±0.71 & 78.53±0.85 & 68.59±1.64 & 64.70±1.74 & 91.06±1.52 & 90.50±1.78 & 77.30±0.74 & 76.46±0.75 & \underline{93.57±0.29} & \underline{92.20±0.33} \\
    \textbf{\MethodName}$\rm _{SAGE}$ & \textbf{81.42±0.68} & \textbf{80.32±0.56} & \underline{69.40±0.82} & \underline{65.71±1.12} & \textbf{92.32±0.70} & \textbf{91.33±0.46} & 77.30±0.81 & 76.53±0.73  &  \textbf{94.33±0.37} & \textbf{92.99±0.49} \\
    \bottomrule
    \end{tabular}}
\caption{Weighted-F1 and Micro-F1 (\% ± standard deviation) of Node Classification on real-world graph datasets. The best results are indicated in \textbf{bold} and the runner-ups are \underline{underlined}. OOM denotes Out-of-Memory.}
  \label{exp_nc_F1}%

\end{table*}%

%% file: tables/exp_ablation.tex
\begin{table}[t]
  \centering
    \resizebox{\linewidth}{!}{
    \begin{tabular}{lcccc}
    \toprule
     \multirow{2}[2]{*}{{\textbf{Method}}}     & \multicolumn{2}{c}{\textbf{Cora}} & \multicolumn{2}{c}{\textbf{Airport}} \\
    \cmidrule(r){2-3} \cmidrule{4-5}
         & AUC    & W-F1    & AUC    & W-F1 \\
    \midrule
    \textbf{\MethodName}  & \textbf{97.91} & \textbf{80.51} & \textbf{97.53} & \textbf{91.75} \\
    \MethodName~(\textit{w/o distortion}) & 97.30  & 79.41  & 97.07  & 91.07  \\
    \MethodName~(\textit{w/o gating}) & 96.96  & 79.17  & 96.86  & 90.87  \\
    \MethodName~(\textit{w/o diverse}) & 94.34  & 79.69  & 94.69  & 90.46  \\
    \MethodName~(\textit{w/o align}) & 97.11  & 79.07  & 96.34  & 90.18  \\
    \bottomrule
    \end{tabular}}
    \caption{Results of ablation study. AUC (\%) for link prediction and Weighted-F1 (\%) for node classification.}
  \label{exp_ablation}%

\end{table}%

%% file: tables/avg_distortion.tex
\begin{table}[t]
  \centering
    \resizebox{\linewidth}{!}{
    \begin{tabular}{lccccc}
    \toprule
     \textbf{Method}     & \textbf{Cora}  & \textbf{Citeseer} & \textbf{Airport} & \textbf{PubMed} & \textbf{Photo} \\
    \midrule
    \textbf{\MethodName} & \textbf{0.22}  & \textbf{0.32}  & \textbf{0.55}  & \textbf{0.21}  & \textbf{0.59}  \\
    MotifRGC & 0.69  & 0.78  & 0.64  & 0.58  & OOM \\
    $\mathcal{Q}$-GCN & 0.80  & 0.91  & 0.70  & 0.75  & 0.78  \\
    \bottomrule
    \end{tabular}}
\caption{Comparison of average embedding distortion. OOM denotes Out-of-Memory.}
  \label{table:avg_distortion}%

\end{table}%

%% file: 6_conclusion.tex
\section{Conclusion}

In this paper, we propose a novel method \MethodName~to address the problem of topological heterogeneity from the perspective of local topology characterization.
We first propose a topology-aware gating mechanism that selects the optimal embedding space for nodes individually by estimating local geometric properties. Then we design diverse Riemannian experts to flexibly construct personalized mixed curvature spaces, effectively embedding the graph into a heterogeneous manifold with varying curvatures at different points. Finally, we present a concise and effective alignment strategy to fairly measure pairwise distances between different embedding spaces. 
Extensive experiments have shown the advantages of \MethodName~on topological heterogeneity. Future work will further explore the potential and broader impact of Riemannian MoE in enhancing the ability of graph foundation models to uniformly handle diverse graph data from a wide variety of types and domains.

%% file: appendix/appendix.tex

\setcounter{figure}{0}
\setcounter{table}{0}
\setcounter{equation}{0}
\renewcommand\thesection{\Alph{section}}
\renewcommand\thefigure{\thesection.\arabic{figure}}
\renewcommand\thetable{\thesection.\arabic{table}}
\renewcommand\theequation{\thesection.\arabic{equation}}

\section{Experiment Details}

\subsection{Datasets Details}
\label{sec:datasets}
\subsubsection{Real-world Datasets.}
We use five real-world datasets, including citation networks, airline networks and co-purchase networks, to evaluate \MethodName~on link prediction and node classification tasks. Statistics of the real-world datasets are concluded in Table \ref{table:datasets}. 


\subsubsection{Synthetic Datasets.}
We synthesize graphs with topological heterogeneity by generating substructures with different topological patterns (e.g., trees, cycles, etc.) and mixing them by adding random edges between them. Considering the extent of topological heterogeneity, we control it by the scale of the synthetic graph, which means that larger graphs contain more complex mixed topologies. For three synthetic graphs Syn.Small, Syn.Middle, and Syn.Large, containing 4027, 10014, and 20018 nodes respectively. We uniformly use the node2vec~\cite{grover2016node2vec} algorithm to initialize the node features of synthetic graphs, where the feature dimension is 128.

\subsubsection{Graph Sectional Curvature.}
\begin{figure*}[htbp]
\centering
\includegraphics[width=1.0\textwidth]{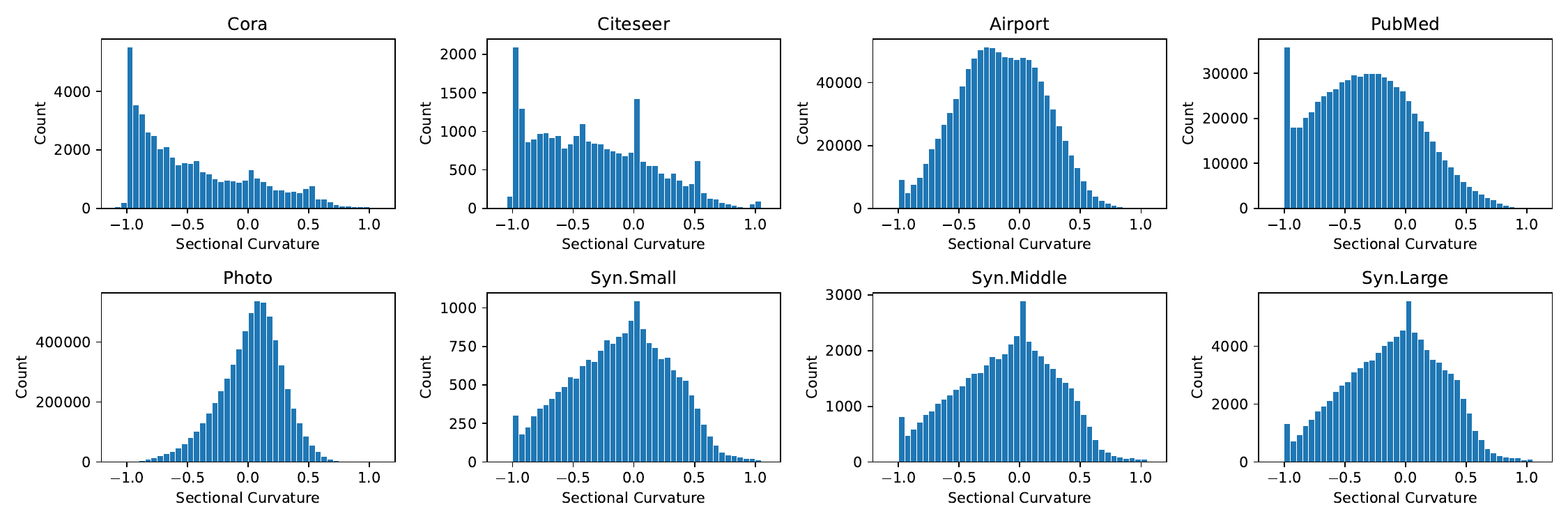} 
\caption{Statistics of sectional curvature distribution.}
\label{fig:SectionalCurvature}
\end{figure*}

We explore topological heterogeneity by analyzing the sectional curvature of the graph. Specifically, according to the parallelogram law~\cite{gu2018learning}, the local topology with positive sectional curvature is more closely matched with cycle-like positively curved space, while the local topology with negative sectional curvature is more closely matched with tree-like negatively curved space. Given a graph $\mathcal{G}$ with the node set $\mathcal{V}$, we can calculate the sectional curvature of the geodesic triangle composed of node $m$ and its neighboring nodes $b$ and $c$ as:
\begin{equation}
\begin{aligned}
\kappa(m ; b, c)=\frac{1}{|\mathcal{V}|} \sum_{a \in \mathcal{V}} g(a, m)^2+\frac{g(b, c)^2}{4} \\ 
-\frac{g(a, b)^2+g(a, c)^2}{2},
\label{eq:SectionalCurvature}
\end{aligned}
\end{equation}
where $g(a,b)$ is the shortest path distance between node $a$ and $b$ on the graph. Figure \ref{fig:SectionalCurvature} shows the sectional curvature statistics of datasets used in this paper, where the degree of near-uniform distribution of positive and negative sectional curvatures is positively correlated with the degree of topological heterogeneity.

\input{tables/datasets}

\subsection{Experiment Setting}
\label{sec:ExperimentSetting}

\subsubsection{Downstream Tasks.}
\label{Sec:Exp_DownstreamTasks}
For the link prediction task, we follow the settings in \cite{chami2019hyperbolic}, spliting the training set, validation set, and test set in proportions of 85\%, 5\%, and 10\%. We sample negative links from any possible links outside the training set (i.e. unobservable links), and keep the number of positive and negative links consistent. We use the Area under the ROC Curve (AUC) and Average Precision (AP) as the evaluation metrics. For node classification tasks, we follow the settings in \cite{fu2021ace,sun2024motif}, and use the Weighted-F1 (W-F1) and Macro-F1 (M-F1) as the evaluation metrics.

\subsubsection{Hyperparameters.}
\label{Sec:Hyperparameters}
Hyperparameters are obtained by grid search, the search range is as follows: the learning rate is \{0.1, 0.01, 0.001\}, the weight decay is \{0.0, 0.0005, 0.001\}, and the loss balance coefficient $\lambda$ is \{1.0, 0.5, 0.1, 0.05, 0.01\}. For all experiments, we train the model for at least 200 epochs, the early stop is set to 100. The other settings are described in Sec 5.1. The parameters of Riemannian experts are optimized by Riemannian Adam~\cite{becigneul2018riemannian}, while the parameters of other models are optimized by Adam~\cite{kingma2014adam}.

\section{Implement Details}
\subsection{Details of \MethodName}
\label{Sec:Details}
In our method, considering the generality across all datasets, the number of Riemannian experts is set to 5, with initial curvatures of \{-3, -1, 0, 1, 3\}. For Cora and Citeseer datasets, the embedding dimension of the Riemannian expert is set to 32, while for other datasets it is set to 16. The local topology sampling employs a simple strategy that utilizes induced subgraphs of ego networks with different radius.

\subsection{Running Environment}
\label{Sec:RunningEnvironment}
We conduct the experiments with:
\begin{itemize}
    \item Operating System: Ubuntu 20.04 LTS.
    \item CPU: Intel(R) Xeon(R) Platinum 8358 CPU@2.60GHz with 1TB DDR4 of Memory.
    \item GPU: NVIDIA Tesla A100 with 40GB of Memory. 
    \item Software: CUDA 11.7, Python 3.8.0, Pytorch 1.13.1, PyTorch Geometric 2.3.1.
\end{itemize}

%% file: tables/datasets.tex
\begin{table}[h]
  \centering
  \resizebox{\linewidth}{!}{
    \begin{tabular}{lrrrr}
    \toprule
    \textbf{Dataset} & \multicolumn{1}{l}{\textbf{\# Nodes}} & \multicolumn{1}{l}{\textbf{\# Edges}} & \multicolumn{1}{l}{\textbf{\# Features}} & \multicolumn{1}{l}{\textbf{\# Classes}} \\
    \midrule
    Cora  & 2,708 & 5,429 & 1,433 & 7 \\
    Citeseer & 3,327 & 4,732 & 3,703 & 6 \\
    PubMed & 19,717 & 44,338 & 500   & 3 \\
    Airport & 3,188 & 18,631 & 4     & 4 \\
    Photo & 7,650 & 119,081 & 745     & 8 \\
    \bottomrule
    \end{tabular}}
    \caption{Statistics of real-world datasets.}
  \label{table:datasets}
\end{table}